\documentclass[letterpaper, 10 pt, conference]{ieeeconf}
\IEEEoverridecommandlockouts                              

\usepackage{times}
\usepackage{epsfig}
\usepackage{graphicx}
\usepackage{amsmath}
\usepackage{amssymb}
\usepackage{graphicx}
\usepackage{amsmath}
\usepackage{booktabs}
\usepackage{algorithm}
\usepackage{algorithmic}
\usepackage{comment}
\usepackage{bm} 

\usepackage[pagebackref=true,breaklinks=true,letterpaper=true,colorlinks,bookmarks=false]{hyperref}

\author{ Jinchang Zhang, Zijun Li, Guoyu Lu
\thanks{ Jinchang Zhang, Zijun Li and Guoyu Lu are with the Intelligent Vision and Sensing (IVS) Lab at  University of Georgia,USA
        {\tt\small guoyulu62@gmail.com}}%
}

\begin{document}

\title{Language-Depth Navigated Thermal and Visible Image Fusion}

\maketitle

\begin{abstract}
Depth-guided multimodal fusion combines depth information from visible and infrared images, significantly enhancing the performance of 3D reconstruction and robotics applications. Existing thermal-visible image fusion mainly focuses on detection tasks, ignoring other critical information such as depth. By addressing the limitations of single modalities in low-light and complex environments, the depth information from fused images not only generates more accurate point cloud data, improving the completeness and precision of 3D reconstruction, but also provides comprehensive scene understanding for robot navigation, localization, and environmental perception. This supports precise recognition and efficient operations in applications such as autonomous driving and rescue missions.
We introduce a text-guided and depth-driven infrared and visible image fusion network. The model consists of an image fusion branch for extracting multi-channel complementary information through a diffusion model, equipped with a text-guided module, and two auxiliary depth estimation branches. The fusion branch uses CLIP to extract semantic information and parameters from depth-enriched image descriptions to guide the diffusion model in extracting multi-channel features and generating fused images. These fused images are then input into the depth estimation branches to calculate depth-driven loss, optimizing the image fusion network. This framework aims to integrate vision-language and depth to directly generate color-fused images from multimodal inputs.

\end{abstract}

\section{Introduction}
Multimodal fusion integrates multi-source information to achieve comprehensive data representation, addressing the limitations of single-modal data \cite{wang2020deep}. Infrared and visible image fusion is a typical task \cite{ma2019infrared}, where infrared imaging is robust in low-light conditions but lacks texture details \cite{kim2021uncertainty}, while visible images contain rich structural and texture information but are affected by lighting variations \cite{li2019illumination}. The complementarity of these modalities allows fused images to retain both thermal targets and texture details, enhancing visual perception \cite{zhu2022clf}. However, existing methods primarily focus on detection-based image fusion, overlooking the role of depth guidance.
Depth-guided multimodal fusion combines depth information from visible and infrared images to improve 3D reconstruction and robotic perception. However, modality differences can cause information loss and feature mismatches. Fusing these modalities enables capturing complete geometric structures in complex environments, improving 3D point cloud quality and model adaptability. Depth-guided fusion is crucial for navigation, localization, and environmental perception, yet achieving comprehensive scene understanding in low-light or obstacle-dense environments remains a significant challenge. For example, in autonomous driving, roads and obstacles must be identified under extreme conditions; in rescue missions, reliable 3D reconstruction is required in environments obscured by smoke or dust.

To address these challenges, we propose a depth-driven visible-infrared image fusion framework based on vision-language models. This framework includes a multi-channel feature extraction module based on diffusion models, a language-guided fusion module, and depth-driven fusion branches.
In the depth-driven module, we design two depth estimation branches for supervised depth estimation of infrared and visible images. The fused image is fed into the trained depth estimation networks, where depth-driven loss is computed by comparing the predicted depth with ground truth, optimizing the image fusion network using depth information.
In the diffusion-based feature extraction module, visible and infrared images are concatenated along the channel dimension to form a four-channel source image, serving as the multi-channel input for the diffusion model and the ground truth for training a self-supervised diffusion model. In the forward process, Gaussian noise is gradually added to the multi-channel data until it becomes nearly pure noise. In the reverse process, a denoising network predicts and removes the noise. After training the noise prediction network, we use it to extract multi-channel features.
Given the lack of constraints from real fused images, we introduce a language-guided fusion process. By generating text descriptions that integrate image and depth information, we comprehensively capture scene details and structure. In the language-driven fusion module, we employ a Depth-Informed Image Captioning Network, which takes infrared and visible images along with their predicted depth as inputs and generates depth-enriched text descriptions. These descriptions are then processed by CLIP’s text encoder to obtain text embedding features.
From the text embeddings, we extract semantic parameters and use an MLP to analyze semantic information. The network outputs semantic information and parameters, which interact with the fusion features through scale adjustment and bias control. Under the guidance of semantic parameters, the fusion process dynamically adjusts feature weights and priorities, producing results that better align with scene requirements.

\begin{figure*}[t]
\begin{center}
\includegraphics[width=17cm, height=7cm]{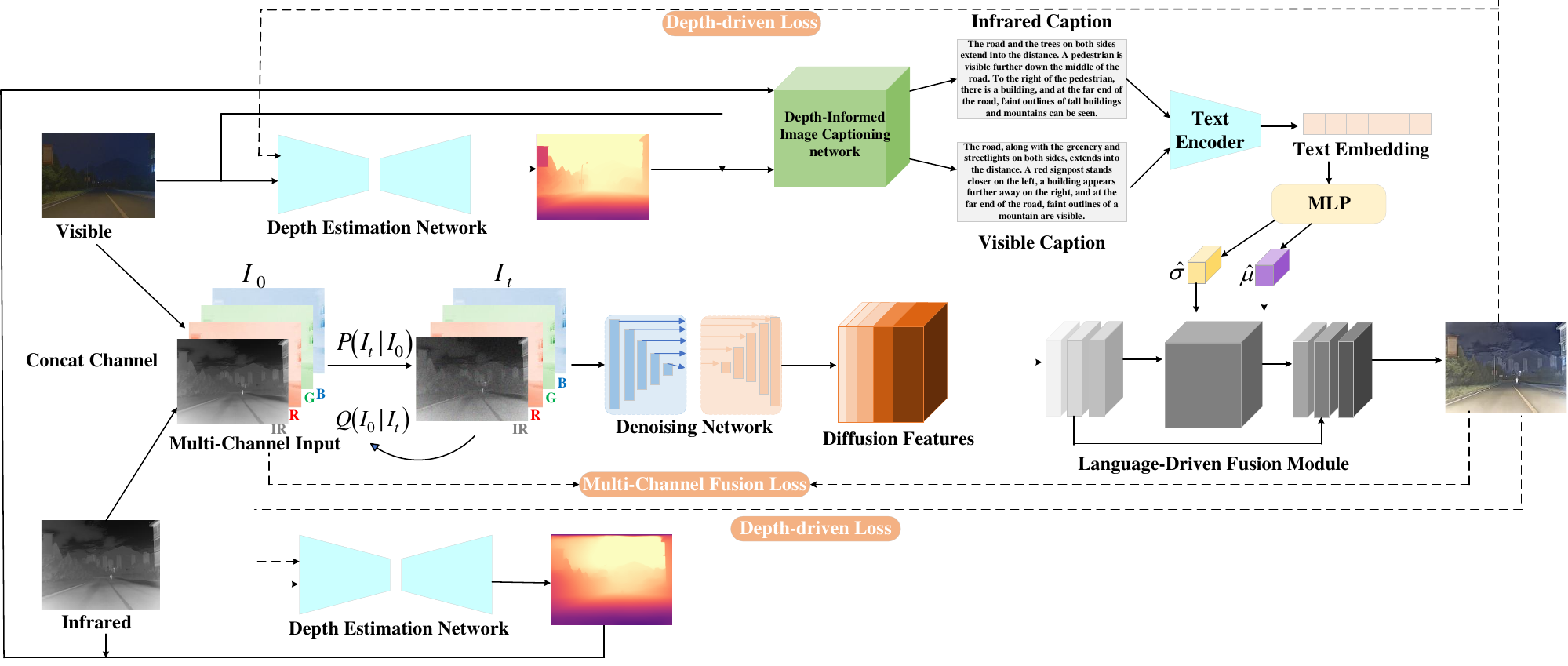}
\end{center}
\vspace{-5 mm}
\caption{Overview of the framework. The visible and infrared images are combined as a four-channel input, serving as ground-truth for self-supervised training of the noise prediction network to extract multi-channel features. In the forward diffusion process, \( I_0 \) and \( I_t \) represent the multi-channel input and the data at timestep \( t \), respectively. \( P(\cdot | \cdot) \) and \( Q(\cdot | \cdot) \) denote the forward and reverse diffusion processes. The multi-channel fusion loss includes intensity loss and gradient loss.
Two depth estimation networks are trained separately for visible and infrared images, generating corresponding depth, which are then input into the depth information description module to produce image-text descriptions containing depth information. Using the CLIP text encoder, text features are extracted, and an MLP predicts semantic information and parameters to guide multi-channel feature reconstruction of fused image. The fused image is processed through the two depth estimation networks, generating depth that are compared with ground-truth to calculate depth-driven loss, optimizing the fusion process.
}
\vspace{-5mm}
\label{overallframework}
\end{figure*}

Overall, our contributions are summarized as follows:
1. We propose a depth-driven, vision-language-based framework for infrared and visible image fusion.
2. We are the first to introduce a depth-driven infrared and visible image fusion method, using depth-driven loss from the fused images to provide spatial information guidance for optimizing the multimodal image fusion network.
3. We introduce a semantic interaction guidance module, which extracts textual semantic information from image content descriptions containing depth information, guiding the multi-channel features extracted by the diffusion model to generate the fused images.
Our framework is shown in Fig \ref{overallframework}.

\section{RELATED WORK}
\subsection{ Conventional Image Fusion Methods}
\label{sec:traditional}
Sparse representation theory represents image signals as a linear combination of the fewest atoms or transformation primitives from an overcomplete dictionary \cite{hamida20183}. In image fusion, it captures data-driven image representations by learning a complete dictionary.
Multi-scale transformation decomposes the original image into sub-images at different scales, mimicking the human visual process and enhancing the visual quality of the fused images \cite{ma2019infrared}. These methods include nonlinear methods, pixel-level weighted averaging, estimation-based methods, and color composite fusion \cite{dogra2017efficient}
Subspace representation-based methods project high-dimensional features into a low-dimensional subspace, capturing the intrinsic structure of the input images while saving processing time and memory \cite{tang2022mdedfusion}. Common methods include PCA \cite{zhang2023joint}, NMF \cite{wang2014fusion}
Saliency detection models simulate human behavior by capturing the most salient regions or objects in an image and have significant applications in computer vision and pattern recognition \cite{wang2020large}. 

\subsection{Deep learning-based Image Fusion Methods}
CNN-based fusion methods, such as PMGI \cite{zhang2020rethinking}, focus on fast image fusion while preserving gradient and intensity. SDNet \cite{zhang2021sdnet} enhances this by adaptively determining the proportion of gradient information to retain, aiming to preserve more detailed textures. 
Diffusion models, a powerful class of deep generative models, have gained prominence in tasks like image generation \cite{sohl2015deep}, inpainting \cite{lugmayr2022repaint}, and image-to-image translation \cite{saharia2022palette}, replacing GANs as the dominant method \cite{croitoru2023diffusion}. These models work by adding Gaussian noise in the forward process and progressively denoising it in the reverse process. They also provide valuable feature representations for discriminative tasks such as classification \cite{zimmermann2021score}, segmentation \cite{baranchuk2021label}, and object detection \cite{chen2023diffusiondet}. Notably, \cite{chen2023diffusiondet} explores the use of diffusion models for visible and infrared image fusion.

\subsection{language-Vision Models}
With advancements in network architectures and large-scale datasets, visual-language models have gained attention in generative modeling. CLIP \cite{radford2021learning}, which aligns image-text pairs using contrastive loss through two neural network encoders, demonstrates strong feature extraction and zero-shot recognition capabilities. Supported by CLIP, various text-driven image generation and processing methods have emerged.
Style-CLIP \cite{patashnik2021styleclip} integrates textual prompts with StyleGAN \cite{karras2020analyzing}, enabling image modifications through text. Beyond GANs, text-conditioned diffusion models like DiffusionCLIP \cite{kim2022diffusionclip} and Stable Diffusion \cite{rombach2022high} have expanded the possibilities for text-guided image generation by combining diffusion models with text encoders and attention mechanisms. These approaches facilitate interactive multimodal fusion and control, offering more flexible image creation and editing options.
TextIF \cite{yi2024text} proposed a text-guided image fusion framework to address the limitations of existing methods in complex scenes, enabling fusion through interactive text inputs. \cite{wang2024infrared} used CLIP to create a language-driven fusion model that links embedding vectors for fusion objectives and image features, along with a language-driven loss function for image fusion.
However, the textual descriptions in these methods are often overly simplistic. To enhance fusion accuracy, we extract enriched textual descriptions from visible and infrared images with their predicted depth, using them as priors to guide the fusion process.

\section{Language-Depth Driven Image Fusion}
Depth estimation provides spatial structural information for image fusion, allowing the fused image to realistically retain key details from both visible and non-visible light. By guiding the fusion process with language, we aim to improve target localization and scene understanding under low-light or occluded conditions, enhancing the robustness of visual perception. Our goal is to construct an infrared and visible image fusion framework driven by both depth estimation and language, fully utilizing the spatial information from depth data and further enhancing the fusion outcome through semantic guidance from language.

\subsection{Depth-driven Image Fusion}
\vspace{-1mm}
Tasks such as 3D reconstruction, robotic navigation, and environmental perception \cite{zhang2024embodiment} often rely on precise depth information from images. However, under low light or challenging exposure conditions, depth estimation may lose crucial details. While infrared imaging is robust to lighting variations, it lacks texture details. By fusing infrared and visible images, we can overcome the limitations of single modalities, allowing depth estimation from fused images to better meet the demands of 3D reconstruction and robotics. Depth information provides essential geometric cues that help align visible and infrared features accurately, ensuring spatial consistency across multimodal inputs. Thus, we aim to leverage depth information to guide image fusion, enhancing the layered structure and detail preservation in the resulting images.
Our depth-driven process is shown in Fig \ref{overallframework}. Given a pair of infrared image \( I_{ir} \in \mathbb{R}^{H \times W \times 1} \) and visible image \( I_{vis} \in \mathbb{R}^{H \times W \times 3} \), we use the ZoeDepth \cite{bhat2023zoedepth} as the depth estimation model, which can be substituted by any other depth estimation models. The infrared and visible images are trained in a supervised manner with the SiLog loss function to learn depth-related information.
For the fused image \( I_F \in \mathbb{R}^{H \times W \times 3} \), it is passed through the trained depth estimation networks, and by comparing the predicted depth with the ground-truth, we calculate the depth-driven loss. This depth information optimizes the image fusion network, enhancing the overall quality of the fused image by improving spatial consistency and structural fidelity.

To enhance the model's robustness to scale variations, we employ the Scale-Invariant Logarithmic Loss (SiLog Loss) \cite{eigen2014depth}, which performs especially well in scenarios with different scales. The SiLog Loss is defined as:
\vspace{-1mm}
\begin{scriptsize}
\begin{equation}
   \mathcal{L}_{\text{SiLog}} = \frac{1}{n} \sum_{i=1}^n \left( \log(y_i) - \log(\hat{y}_i) \right)^2 - \frac{1}{n^2} \left( \sum_{i=1}^n \left( \log(y_i) - \log(\hat{y}_i) \right) \right)^2 
\end{equation}
\vspace{-1mm}
\end{scriptsize}

\noindent where \( y_i \) denotes the ground-truth,
\( \hat{y}_i \) denotes the predicted depth, \( n \) is the number of pixels in image.

\subsection{Multi-channel Diffusion Feature Extraction Based on Diffusion Models}
This section provides a detailed description of the language-guided diffusion model for image fusion.
In multimodal image fusion tasks, the key objective is to extract complementary information between modalities while preserving details and structural information. The significant differences between infrared and visible images make it challenging for deep learning-based feature extraction methods to fully utilize their complementary characteristics. Diffusion models, through the process of forward noise addition and reverse denoising, can effectively model subtle correlations in high-dimensional feature spaces. By concatenating infrared and visible images into a four-channel input, diffusion models can jointly model multimodal features in the multi-channel space, generating high-quality fused features. This approach maximizes the utilization of multimodal information and enhances fusion performance.
The visible and infrared image pairs are concatenated along the channel dimension to form a multi-channel input for the diffusion model. In the forward process, Gaussian noise is gradually added to the multi-channel data until it approaches pure noise (e.g.,\( P(I_t | I_{t-1}) \) ). In the reverse process, a denoising network (e.g., \( Q(I_{t-1} | I_t) \)) predicts and removes the added noise.
We use the depth estimation network trained in the previous section to generate depth for both the visible and infrared images. These images and their corresponding depth are input into an image description network, which generates text descriptions with depth information based on the input images and depth, followed by CLIP extraction of text features. The diffusion features extracted from the diffusion model, along with the text features, are input into a semantic interaction network to produce the fused image, and the loss is calculated with the ground-truth using the depth estimation network trained on visible and infrared images.
Given a pair of infrared images \( I_{\text{ir}} \in \mathbb{R}^{H \times W \times 1} \) and visible images \( I_{\text{vis}} \in \mathbb{R}^{H \times W \times 3} \), where \( H \) and \( W \) represent the height and width, respectively, they capture different types of visual information. Infrared images provide clear thermal radiation information in low-light or complex environments, while visible images present rich colors and details. By concatenating the infrared and visible images, a 4-channel image is formed, denoted as \( I_{\text{f}} \in \mathbb{R}^{H \times W \times 4} \). Learning the joint potential structure of multi-channel data enables better integration of these two modalities, thereby retaining and enhancing useful features in the fused image. We employ the diffusion process proposed in the Denoising Diffusion Probabilistic Model \cite{ho2020denoising} to model the distribution of multi-channel data. The forward diffusion process of multi-channel images is conducted by gradually adding noise over \( T \) time steps. In the reverse process, the noise is gradually removed over \( T \) time steps, learning the joint latent structure of infrared and visible images. The learning process of the joint potential structure helps the model understand the relationships and differences between infrared and visible images, thereby enabling better balance and selection of important information from each modality during the fusion process. This contributes to generating more consistent, and high-quality fused images, avoiding information loss or inconsistencies. For the update formulas of the Forward Diffusion Process and Reverse Diffusion Process, we refer \cite{yue2023dif}.

Forward Diffusion Process:
The forward diffusion process can be viewed as a Markov chain that gradually adds Gaussian noise to the data with $T$ time steps. For the first time step, the noisy image $\bm{I_{1}}$ can be formulated as \ref{x1} and at timestep $t$, the noisy image $\bm{I_{t}}$ can be represented as \ref{xt}:
\begin{eqnarray}
	\bm{I_{1}}=\sqrt{\alpha_{1}}\bm{I_{f}}+\sqrt{1-\alpha_{1}}\bm{\gamma}
	\label{x1}
    \vspace{-5mm}
\end{eqnarray}
\vspace{-7mm}
\begin{eqnarray}
P(\bm{I_{t}}|\bm{I_{t-1}})=\mathcal{N}(\bm{I_{t}};\sqrt{\alpha_{t}}\bm{I_{t-1}},(1-\alpha_{t})\bm{Z})
	\label{xt}
\end{eqnarray}

\noindent where $\bm{I_{f}}$ is original input $\bm{I_{f}}\in \mathbb{R}^{H×W×4}$, $\bm{\gamma}$ is the Gaussian noise. $\alpha_{t}$ is the variance schedule that controls the variance of the Gaussian noise added in time step $t$, and $\bm{Z}$ denotes the standard normal distribution. $\bm{I_{t}}$ and $\bm{I_{t-1}}$ represents the noisy images generated for $t$ and $t-1$ times, respectively. 
By applying Equations \ref{x1} and \ref{xt}, the relationship between $\bm{I_{t}}$ and $\bm{I_{f}}$ can be derived as follows:
\begin{eqnarray}
P(\bm{I_{t}}|\bm{I_{f}})=\mathcal{N}(\bm{I_{t}};\sqrt{\bar{\alpha}_{t}}\bm{I_{f}},(1-\bar{\alpha}_{t})\bm{Z})
	\label{x0xt}
    \vspace{-4mm}
\end{eqnarray}

Reverse Diffusion Process:
The reverse diffusion process obtains the original multi-channel image through denoising. In each timestep of the reverse process, the denoising operation is performed on the noisy multi-channel image $\bm{I_{t}}$ to obtain the previous image $\bm{I_{t-1}}$. The probability distribution of $\bm{I_{t-1}}$ under the condition $\bm{I_{t}}$ can be formulated as:
\vspace{-1mm}
\begin{eqnarray}
	Q(\bm{I_{t-1}}|\bm{I_{t}})=\mathcal{N}(\bm{I_{t-1}};\mu_{\theta}(\bm{I_{t}},t),\sigma_{t}^{2}\bm{Z})
	\label{xtxt-1}
\end{eqnarray}
\noindent where $\sigma_{t}^{2}$ is the variance of the conditional distribution $Q(\bm{I_{t-1}}|\bm{I_{t}})$, which can be formulated as:
\vspace{-2mm}
\begin{eqnarray}
    \sigma_{t}^{2} &=& \frac{1-\bar{\alpha}_{t-1}}{1-\bar{\alpha}_{t}} (1-\alpha_{t})
    \label{variance}
\end{eqnarray}

The mean $\mu_{\theta}(\bm{I_{t}},t)$ of the conditional distribution $Q(\bm{I_{t-1}}|\bm{I_{t}})$ can be formulated as:
\vspace{-3mm}
\begin{eqnarray}
	\mu_{\theta}(\bm{I_{t}},t)=\frac{1}{\sqrt{\alpha_{t}}}(\bm{I_{t}}-\frac{\beta_{t}}{\sqrt{1-\bar{\alpha}_{t}}}\epsilon_{\theta}(\bm{I_{t}},t))
	\label{mean}
\end{eqnarray}
\noindent where $\epsilon_{\theta}(\cdot,\cdot)$ is the denoising network. The inputs of $\epsilon_{\theta}(\cdot,\cdot)$ are the timestep $t$ and the noisy multi-channel image $\bm{I_{t}}$.

Loss Function of Diffusion Process:
We sample a pair of visible and infrared image pairs $(\bm{I_{ir}}, \bm{I_{vis}})$ to form the multi-channel image $\bm{I}$. Then we sample the noise $\bm{\gamma}$ from the standard normal distribution. Third, we sample the timestep $t\sim U(\{1,...,T\})$ from the uniform distribution. The loss function is optimized by calculating the difference between the noise predicted by the model and the actual noise added, which is used to guide the training procedures of the model.
\begin{eqnarray}
	\mathcal{L}_{diff}=\left \| \bm{\gamma}-\epsilon_{\theta}(\sqrt{\bar{\alpha}_{t}}\bm{I_{0}}+\sqrt{1-\bar{\alpha}_{t}}\bm{\gamma},t)\right \|_{2}
	\label{mean}
\end{eqnarray}

\textbf{Structure of the Denoising Network :}
In order to predict the noise added in the forward diffusion process, the structure of the denoising network adopts the U-Net structure used in SR3 \cite{saharia2022image}.

\textbf{Loss Function of Fusion Process: }
To preserve sufficient texture information in the final fused image, we apply a gradient loss to maintain gradient fidelity. To directly generate a three-channel fused image while preserving gradients, we used the multi-channel gradient loss $\mathcal{L}_{MCG}$ \cite{yue2023dif}, which is defined as follows:
\vspace{-2mm}
\begin{eqnarray}
	\mathcal{L}_{MCG}=\frac{1}{HW}\sum_{i=1}^3\left \| \nabla \bm{I_{f}^{i}} - \mbox{max}(\nabla \lvert \bm{I_{ir}} \lvert,\nabla \lvert \bm{I_{vis}^{i}} \lvert ) \right \|_{1}
	\label{spatialloss2}
\end{eqnarray}
\noindent where $\nabla$ represents the gradient operator. $\bm{I_{f}^{1}}$, $\bm{I_{f}^{2}}$ and $\bm{I_{f}^{3}}$ represent the three channels (i.e., red, green and blue) of the fused image $\bm{I_{f}}$. $\bm{I_{vis}^{1}}$, $\bm{I_{vis}^{2}}$ and $\bm{I_{vis}^{3}}$ denote the three channels of the input visible image $\bm{I_{vis}}$.
We apply intensity loss to make the fused image have an intensity distribution similar to the infrared image and the visible image.  We use multi-channel intensity loss $\mathcal{L}_{MCI}$ \cite{yue2023dif}, which is:
\vspace{-2mm}
\begin{eqnarray}
	\mathcal{L}_{MCI}=\frac{1}{HW}\sum_{i=1}^3\left \| \bm{I_{f}^{i}}- \mbox{max}(\bm{I_{ir}},\bm{I_{vis}^{i}}) \right \|_{1}
	\label{spatialloss2}
    \vspace{-3mm}
\end{eqnarray}
We directly generate three-channel fused images with multi-channel gradient and intensity losses. The final loss $\mathcal{L}_{f}$ can be formulated as
$\mathcal{L}_{f}=\mathcal{L}_{MCG}+\mathcal{L}_{MCI}$

\subsection{Language-driven Image Fusion}
In the process of image fusion, deep learning often relies on multiple loss functions. However, due to the lack of constraints from a real fused image, it becomes challenging to regulate the fusion output through loss functions, and many problems cannot be explicitly modeled, limiting the model's performance. We propose a language-guided approach to the fusion process. While text descriptions based on images can provide visual features, they lack geometric constraints and fail to capture the spatial layout and depth relationships of the scene. To address this, we propose generating text descriptions that combine image and depth information, offering a more comprehensive representation of scene details and structure.
We utilize \cite{mahmoud2023enhancing} as our Depth-Informed Image Captioning Network, where the input consists of an image and its corresponding depth, and the output is a text description enriched with depth information. Given a pair of infrared image \( I_{ir} \) and visible image \( I_{vis} \), along with their corresponding depth predicted by a depth estimation network, these inputs are processed through the Depth-Informed Image Captioning Network. The resulting text descriptions are concatenated and passed through the frozen text encoder of CLIP to obtain text embedding features.
To utilize text embedding features to guide image fusion, we extract semantic parameters from the text embeddings. These parameters contain high-level guiding information such as object attributes, spatial relationships, and depth information within the scene. We use an MLP to explore these relationships and further map the text semantic information \(\hat{\sigma}\) and semantic parameters \(\hat{\mu}\), with the fusion features represented as \(F_i\). The semantic parameters guide the fusion features using the formula 
$  \hat{F}_f^i = (1 + \hat{\sigma}) \odot F_f^i + \hat{\mu}$,
where \(\odot\) denotes the Hadamard product.
These parameters interact with the fusion features through feature modulation, which includes scale adjustment and bias control, adjusting feature expressions from different perspectives. This effectively supplements contextual information, enhances feature representation, and ensures alignment of multi-modal features in the semantic space. For the image fusion network, residual connections are employed to reduce the difficulty of network fitting. Under the guidance of semantic parameters, the fusion process dynamically adjusts feature weights and priorities, producing results that better meet scene requirements. The final output is a fused image with rich hierarchical structure, clear semantics, and comprehensive information.

\section{Experiment}

\begin{figure}[t]
\begin{center}
\includegraphics[width=8cm, height=4cm]{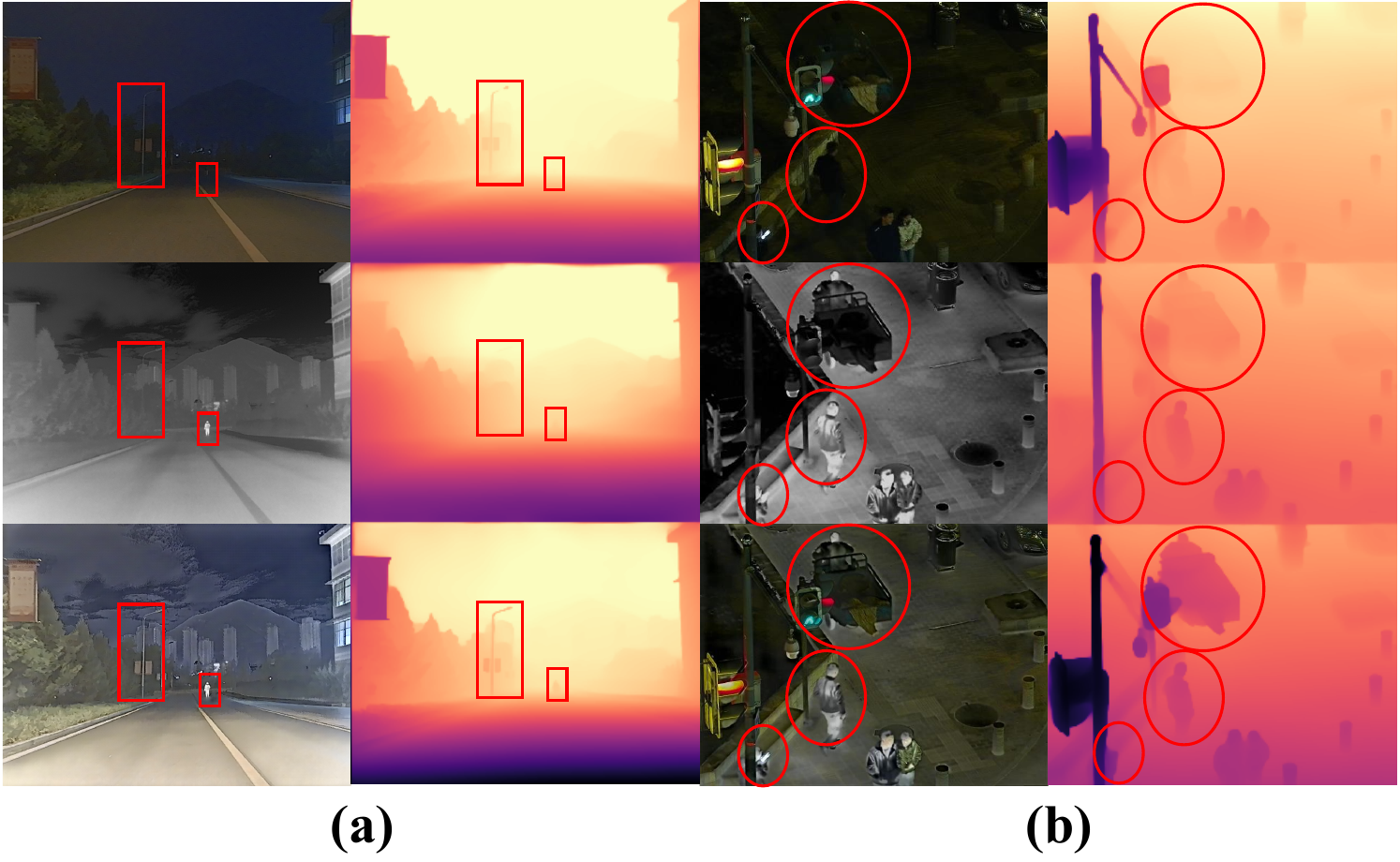}
\end{center}
\vspace{-6 mm}
\caption{Comparison of Depth Estimation for Visible, Infrared, and Fused Images: The visible, infrared, and fused images of images a and b, along with their corresponding depth, are arranged in sequence to visually demonstrate the depth estimation results across different modalities.}
\label{whydepth}
\vspace{-3 mm}
\end{figure}

\begin{table}[t]
\centering
\resizebox{0.47\textwidth}{!}{
\begin{tabular}{lccccccccc}
\toprule
\textbf{Method} & \textbf{SF}  & \textbf{Qab/f}  & \textbf{MI} & \textbf{SD} & \textbf{VIF} \\
\midrule
DeepFuse \cite{ram2017deepfuse}  & 8.3500            & 0.3847                    & 13.2205     & 66.8872    & 0.5752       \\
DenseFuse \cite{li2018densefuse}  & 9.3238           & 0.4735                  & 13.7053     & 81.7283    & 0.6875       \\
RFN-Nest \cite{li2021rfn}   & 5.8457            & 0.3292                  & 13.4547     & 67.8765    & 0.5404       \\
PMGI \cite{li2021rfn}       & 8.7195            & 0.3787                  & 13.7376     & 69.2364    & 0.6904       \\
U2Fusion \cite{xu2020u2fusion}   & 11.0368           & 0.3934                  & 13.4453     & 66.5035    & 0.7680       \\
IFCNN \cite{zhang2020ifcnn}     & 11.8590          & 0.4962                & 13.2909     & 73.7053    & 0.6090       \\
FusionGAN \cite{ma2019fusiongan}  & 8.0476           & 0.2682                  & 13.0817     & 61.6339    & 0.4928       \\
MEFGAN \cite{xu2020mef}     & 7.8481            & 0.2076               & 13.9454     & 43.7332    & 0.7330       \\
SeAFusion \cite{tang2022image} & 11.9355  & 0.4908 & 14.0663 & 93.3851 & 0.8919 \\
YDTR \cite{tang2022ydtr}      & 3.2567           & 0.1410                   & 12.3865     & 56.0668    & 0.2792       \\
MATR \cite{tang2022matr}       & 5.3632           & 0.2723                & 13.0705     & 78.0720    & 0.3920       \\
UMF-CMGR \cite{wang2022unsupervised}   & 8.2388           & 0.3671                   & 12.6301     & 60.7236    & 0.3934       \\
TGFuse \cite{rao2023tgfuse}          & 11.3149 & 0.5863 & 13.9676  & 94.7203 & 0.7746 \\
ours       & 13.3423 & 0.6213 & 16.8924  & 92.1103 & 0.9012 \\
\bottomrule 
\end{tabular}}
\vspace{-2mm}
\caption{Quantitative evaluation results on the TNO dataset.}
\vspace{-10mm}
\label{tro}
\end{table}

\subsection{Implementation Details and Datasets}
\textbf{Dataset:} 
We conducted experiments on three public datasets: LLVIP \cite{jia2021llvip}, RoadScene \cite{xu2020u2fusion}, and TNO \cite{toet2017multiband}. We trained our depth estimation model using the KAIST dataset \cite{hwang2015multispectral} because it includes visible light, infrared images, and LiDAR measurements, with the LiDAR data serving as the ground-truth for our depth estimation model. Additionally, the KAIST dataset covers various scenes, lighting conditions, and weather scenarios. The LLVIP, TNO, and RoadScene datasets were used to evaluate the performance of our proposed method.



\begin{figure}[t]
\begin{center}
\includegraphics[width=8.5cm, height=5cm]{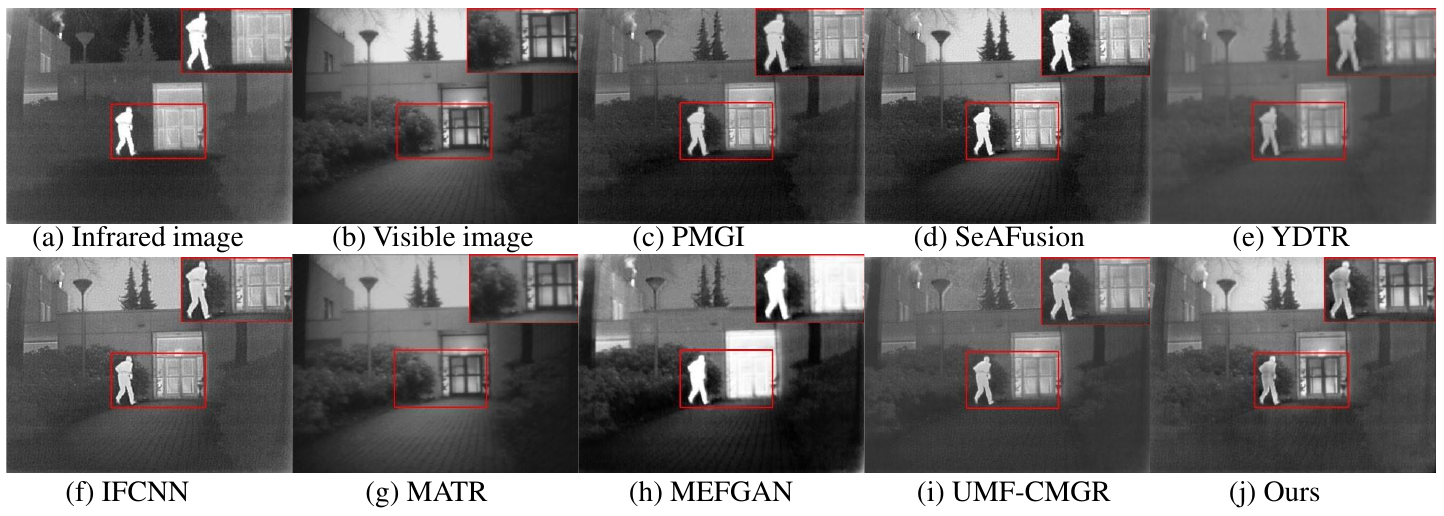}
\end{center}
\vspace{-6 mm}
\caption{Infrared and visible image fusion experiment on ``human” images }
\label{house}
\vspace{-8 mm}
\end{figure}

\subsection{Evaluation of Depth Estimation Performance of Different Modalities}
In this experiment, we performed depth estimation on visible light, infrared, and fused images. The results show that in nighttime environments, visible light images fail to detect certain objects, leading to missing areas in the corresponding depth. Infrared images, on the other hand, can capture objects that are undetectable in visible light under low-light conditions, providing better depth coverage, but also have the drawback that some objects may not appear due to a lack of thermal radiation.
The fused image combines the advantages of both visible light and infrared images, enabling the detection of objects that each individual modality might miss, thus providing more depth information, particularly in complex environments. The experimental results show that the fused depth clearly presents more object details and improves depth estimation performance under low-light or poor imaging conditions.
In Fig. \ref{whydepth}(a), the visible light image shows the streetlamp pole and sign clearly, but pedestrians and distant mountains appear blurry, and the depth only displays the streetlamp pole and sign. The infrared image, however, captures pedestrians and distant mountains, but misses the streetlamp pole and sign. The fused image displays all objects clearly, and the corresponding depth illustrates the complementary effects of both modalities.
In Fig. \ref{whydepth}(b), the visible light image clearly shows the traffic light, but roadside vehicles and pedestrians appear blurry and are not represented in the depth. The infrared image captures these objects clearly, but the top of the traffic light is not visible. The fused image accurately presents all objects and their details, and the depth clearly outlines pedestrians and vehicles, with the top of the traffic light also displayed.

\vspace{-1mm}
\subsection{Comparison with State-of-the-art Fusion Methods}
\vspace{-1mm}
In Fig. \ref{house}, the pedestrian in the infrared image represents salient information, while the red-boxed area in the visible image serves as background. We present fusion results of different methods, highlighting key differences with red boxes. Other models achieve information fusion to some extent but struggle to highlight both the salient infrared information and the low-noise visible background.
In Table \ref{tro}, we compared all methods on five evaluation indicators. Our method achieved the best results on four and ranked third on another.
In Table \ref{vip}, our method was compared with the latest approaches on the LLVIP dataset, achieving the best results on five indicators while maintaining excellent performance. Through visualization and metric evaluation, our method demonstrates a significant performance advantage.

\begin{table}[t]
\centering
\resizebox{0.47\textwidth}{!}{
\begin{tabular}{lccccccccc}
\toprule
\textbf{Method} & \textbf{SF}  & \textbf{Qab/f} & \textbf{MI} & \textbf{SD} & \textbf{VIF} \\
\midrule
DeepFuse \cite{ram2017deepfuse}  & 12.4175        & 0.4620            & 14.0444 &0.4586 &38.3328      \\
DenseFuse \cite{li2018densefuse}  & 12.5900            & 0.4700                  & 14.0723 & 0.4669 & 38.7011       \\
RFN-Nest \cite{li2021rfn}    & 10.6825            & 0.3844                & 14.1284 & 0.4658&  39.7194      \\
PMGI \cite{li2021rfn}       & 12.0997         & 0.3951               & 14.0737 &0.4487 &37.9572       \\
U2Fusion \cite{xu2020u2fusion}   & 17.2889         & 0.4985         & 13.4141  &0.4917  &37.4284       \\
IFCNN \cite{zhang2020ifcnn}      & 21.7698          & 0.6092               & 14.4835 &0.6762 &44.0938       \\
FusionGAN \cite{ma2019fusiongan}  & 9.2062          & 0.0600          & 12.8981 & 0.1141 & 26.9133      \\
MEFGAN \cite{xu2020mef}     & 15.1905           & 0.3644               & 13.9575& 0.8720 &59.7947      \\
SeAFusion \cite{tang2022image}   & {20.9194}  & {0.6181}& 14.9016 & 0.8392&  51.8096
 \\
YDTR \cite{tang2022ydtr}       & 7.0755           & 0.1961                   & 13.3858 &0.3365& 33.1625     \\
MATR \cite{tang2022matr}       & 13.5066           & 0.4282                  & 11.9989 &0.4575& 34.1515
     \\
UMF-CMGR \cite{wang2022unsupervised}  & 13.4481         & 0.3707                 & 13.4037 &0.3841 &35.1731
    \\
TGFuse  \cite{rao2023tgfuse}          & 21.9161  &0.6535  &14.0007 & 0.8737  &43.6293 \\
ours            & 23.4262  &0.6980  &15.2352 & 0.9420  &54.1251 \\
\bottomrule
\end{tabular}}
\vspace{-1mm}
\caption{Quantitative evaluation results on the LLVIP dataset.}
\vspace{-9mm}
\label{vip}
\end{table}

\begin{figure}[t]
\begin{center}
\includegraphics[width=8.5cm, height=9cm]{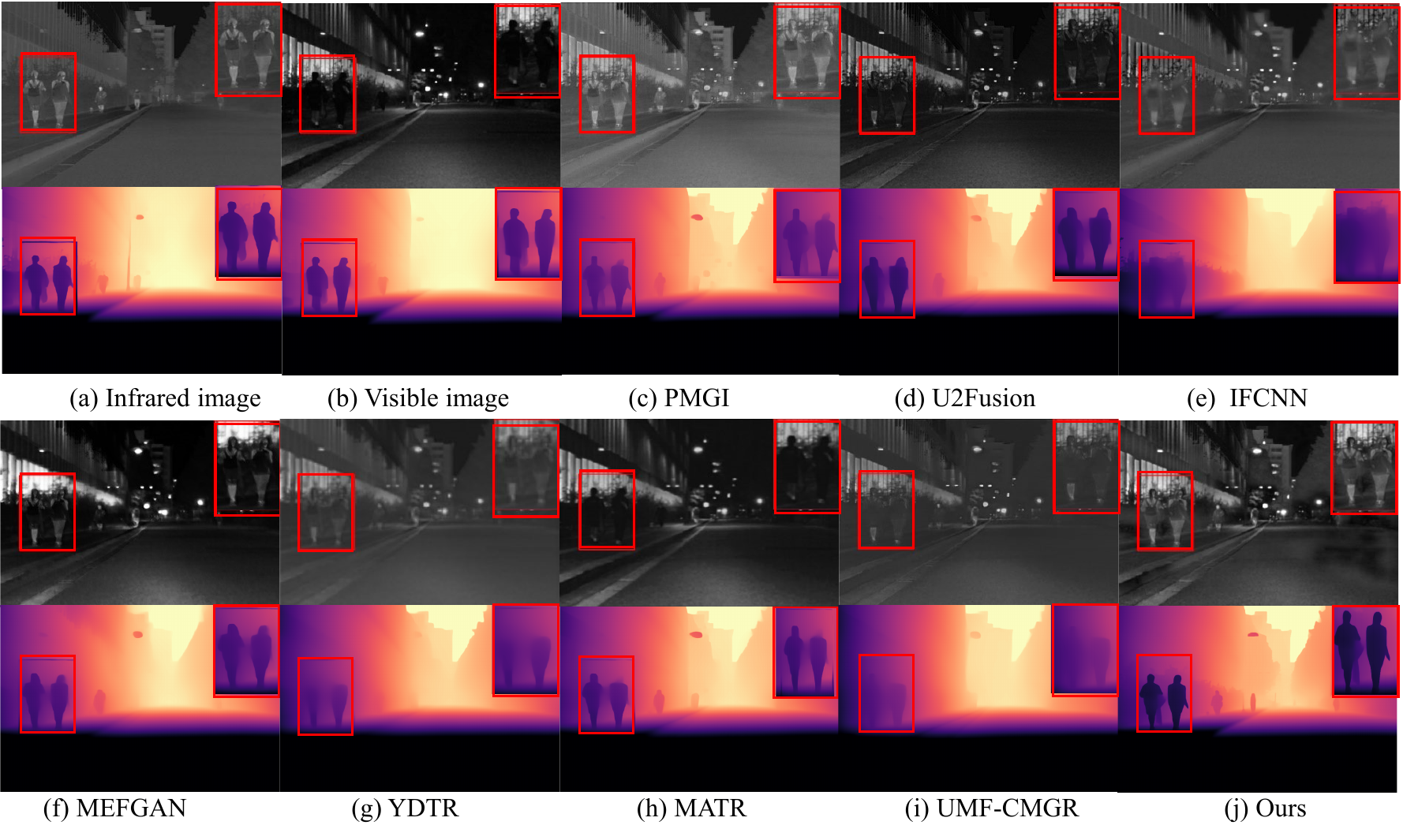}
\end{center}
\vspace{-7 mm}
\caption{Experiments on infrared and visible image fusion and estimated depth on ``street" images. }
\label{compdepth}
\vspace{-9 mm}
\end{figure}

\subsection{Evaluation of Fusion Models Through Depth-Aware Analysis}
Figure \ref{compdepth} shows the fusion results of visible and infrared images from different models, along with their corresponding depth. In the infrared image, distant buildings and utility poles are visible, but the buildings are not clearly represented in the depth, likely due to their subtle features. In the visible image, distant buildings are not visible, and due to the blur in the infrared image, the corresponding depth lacks clear representation of the people.
For the fusion results of different models, both people and distant buildings are generally visible in the fused images. However, the clarity of the people varies significantly, and this difference is even more pronounced in the depth. For example, the depth generated by models e, g, and i show very blurred representations of people. In contrast, our model produces the clearest depiction of people in the depth, even surpassing the clarity seen in the original infrared and visible images. Additionally, distant buildings are represented with exceptional clarity, highlighting the superior performance of our fusion approach.
In Table \ref{depth}, we further compare the impact of different depth estimation models on the fusion results. Our findings suggest that more accurate depth estimation models guide the fusion process to produce better results. However, the performance differences across models were not substantial, indicating that the introduction of depth information itself effectively enhances the fusion quality. This demonstrates the robustness of our approach, where even moderate depth estimation models can significantly contribute to improved fusion outcomes.

\begin{table}[t]
\centering
\resizebox{0.48\textwidth}{!}{%
\begin{tabular}{lccccc}
\hline
\textbf{Method}       & \textbf{SF}        & \textbf{Qab/f}      & \textbf{MI}        & \textbf{SD}        & \textbf{VIF}       \\ \hline

SDC-Depth\cite{wang2020sdc}            & 22.4213  &0.6721  &14.51421 & 0.8983  &52.8965         \\
Adabins \cite{bhat2021adabins}           & 22.9504  &0.6844 &14.7840 & 0.9092  &53.2134         \\
DepthFormer \cite{li2023depthformer}         & 22.9842 &0.6912  &15.0213 & 0.9229  &53.6234            \\
MiM (large) \cite{xie2023revealing}               & 23.1261  &0.6980  &15.2352 & 0.9420  &54.1251   \\
\hline
\end{tabular}}
\vspace{-1mm}
\caption{Quantitative Evaluation Results on Road Scene Dataset. }
\vspace{-6mm}
\label{depth}
\end{table}

\begin{table}[t]
\centering
\resizebox{0.48\textwidth}{!}{
\begin{tabular}{lllllllll}
\multicolumn{1}{c}{Diff} & 
\multicolumn{1}{c}{Dep} & 
\multicolumn{1}{c}{Lan} &  
\multicolumn{1}{c}{SF} & 
\multicolumn{1}{c}{Qab/f} & 
\multicolumn{1}{c}{MI} & 
\multicolumn{1}{c}{SD} & 
\multicolumn{1}{c}{VIF} & 

\\ \hline 
  &  & & 7.6540 & 0.4479 & 10.3591  & 61.4989 & 0.6811   
\\  \hline 
\checkmark    &  &  & 10.9842 & 0.5293 & 13.5140  & 79.4214 & 0.8123    
\\  \hline 
\checkmark   &\checkmark &   & 11.5870 & 0.5744 & 14.7423  & 84.2301 & 0.8549 
\\  \hline  
\checkmark   &\checkmark & \checkmark & 13.3423 & 0.6213 & 16.8924  & 92.1103 & 0.9012 

\\  \hline  
\end{tabular}}
\vspace{-1mm}
\caption{Ablation study of our methods on the TNo data: Diff: Diffusion Process. Dep: Depth-driven Module. Lan: Language-Driven Fusion Module. }
\vspace{-6mm}
\label{Ablation}
\end{table}
\vspace{-3mm}

\subsection{Ablation Study}

\textbf{Depth-driven Module: } 
To verify the effectiveness of the depth-driven module, we removed its loss calculation step while keeping other components unchanged. As shown in rows 2 and 3 of Table \ref{Ablation}, all evaluation metrics decreased after removal. The depth supervision module provides crucial geometric and structural information, enhancing the model’s understanding of scene layout. Without it, the model lacks depth constraints during fusion, impacting performance.

\textbf{Diffusion Process: } We use the diffusion model to extract multi-channel fusion features from visible and infrared light. To verify the effectiveness of the diffusion model, we replaced it with an autoencoder network while keeping the architecture unchanged. The results in rows 1 and 2 of Table \ref{Ablation} show that after removing the diffusion process, the model's performance decreases across various metrics, indicating that the diffusion model effectively extracts complementary multi-channel information.

\textbf{Language-Driven Fusion Module :}
To verify the effectiveness of the language guidance module, we set the MLP-predicted parameters to fixed values of 1 and 0, thereby removing the influence of language on image fusion. Observing rows 3 and 4 in Table \ref{Ablation}, we find a significant performance drop, indicating that text descriptions containing depth information can provide prior for image fusion.
\vspace{-3mm}
\section{Conclusion}
In this paper, we propose a language-guided, depth-driven fusion network for infrared and visible images. The network includes an image fusion branch and two depth estimation branches. The image fusion branch extracts multi-channel complementary information using a diffusion model and incorporates a text-guided module. By leveraging CLIP, we extract semantic information and parameters from image descriptions with depth details to guide the diffusion model in feature extraction and fused image generation. The fused images are then processed by the depth estimation branches, which compute depth-driven loss to optimize the fusion network. Our approach explores a depth-driven, vision-language-based fusion framework aimed at directly generating color-fused images from multi-modal inputs, improving depth estimation in complex lighting conditions.

\bibliographystyle{ieee_fullname}
\bibliography{egbib}

\end{document}